\documentclass{article}

\pdfoutput=1
\usepackage{PRIMEarxiv}

\usepackage[utf8]{inputenc} % allow utf-8 input
\usepackage[T1]{fontenc}    % use 8-bit T1 fonts
\usepackage{hyperref}       % hyperlinks
\usepackage{url}            % simple URL typesetting
\usepackage{booktabs}       % professional-quality tables
\usepackage{amsfonts}       % blackboard math symbols
\usepackage{nicefrac}       % compact symbols for 1/2, etc.
\usepackage{microtype}      % microtypography
\usepackage{lipsum}
\usepackage{fancyhdr}       % header
\usepackage{graphicx}       % graphics
\graphicspath{{media/}}     % organize your images and other figures under media/ folder
\usepackage{amssymb}
\usepackage{amsmath}
\usepackage{subfig}
\usepackage{bm}
\usepackage{xcolor}
\usepackage{setspace}
\title{Gaussian Process Model with Tensorial Inputs and Its Application to the Design of 3D Printed Antennas
\thanks{\textit{\underline{Citation}}: 
\textbf{Authors. Title. Pages.... DOI:000000/11111.}} 
}

\author{
  Xi Chen, Yashika Sharma, Hao Helen Zhang, Xin Hao, Qiang Zhou \\
  The University of Arizona \\
  Tucson, AZ, USA\\
  \texttt{\{xic, yashikasharma, hxin, zhouq\}@arizona.edu} \\
  \texttt{hzhang@math.arizona.edu} \\
}

\begin{document}
\maketitle

\begin{abstract}
In simulation-based engineering design with time-consuming simulators, Gaussian process (GP) models are widely used as fast emulators to speed up the design optimization process. In its most commonly used form, the input of GP is a simple list of design parameters. With rapid development of additive manufacturing (also known as 3D printing), design inputs with 2D/3D spatial information become prevalent in some applications, for example, neighboring relations between pixels/voxels and material distributions in heterogeneous materials. Such spatial information, vital to 3D printed designs, is hard to incorporate into existing GP models with common kernels such as squared exponential or Matérn. In this work, we propose to embed a generalized distance measure into a GP kernel, offering a novel and convenient technique to incorporate spatial information from freeform 3D printed designs into the GP framework. The proposed method allows complex design problems for 3D printed objects to take advantage of a plethora of tools available from the GP surrogate-based simulation optimization such as designed experiments and GP-based optimizations including Bayesian optimization. We investigate the properties of the proposed method and illustrate its performance by several numerical examples of 3D printed antennas. The dataset is publicly available at: \url{https://github.com/xichennn/GP_dataset}.
\end{abstract}

% keywords can be removed
% \keywords{First keyword \and Second keyword \and More}

\section{Introduction}
\label{sec:intro}

%% For citations use: 
%%       \cite{<label>} ==> Jones et al. [21]
%%       \cite{<label>} ==> [21]
%%

Computer models are widely used in engineering areas to study and predict the behavior of complex systems. With the ever-increasing complexity of such models, the simulation time, however, becomes prohibitive that it’s impractical to achieve the conventional objectives, e.g. simulation-based optimization, uncertainty assessment, sensitivity analysis, etc., within a feasible time window \cite{conti2010bayesian,fricker2013multivariate}.  As a remedy of this, metamodels (aka. surrogates, emulators) which statistically approximate the computer models hence more time-efficient, have become prevalent \cite{koziel2013surrogate}. Gaussian processes (GPs) \cite{rasmussen2003gaussian,santner2003design}, among all the metamodeling techniques, stand out due to its two distinctive advantages: the ability to quantify prediction uncertainty and requirements of a relatively small number of sample data points for model training. Computer simulations can be categorized as deterministic and stochastic. In this paper, we focus on deterministic simulations where identical outputs will be generated for the same inputs.

GPs are characterized by their kernel functions, which quantify the similarity between pairs of data inputs. For example, many traditional kernels rely on Euclidean distance metrics, such as the radial basis function (RBF). However, with rapid advancements in additive manufacturing, or 3D printing, technologies, the landscape of 2D/3D free-form designs has become increasingly intricate, involving spatially varying material properties. Tensors have emerged as a versatile representation for such data, encapsulating complex structures \cite{wang2022bayesian}. While a straightforward approach for handling tensorial inputs involves flattening them into vectors and subsequently applying Euclidean-distance based kernel functions, this method presents serious drawbacks. The flattening process and the subsequent Euclidean distance measure neglects the nuanced structural information inherent in the data, leading to suboptimal designs. Consequently, there is a strong need for more sophisticated approaches that can effectively leverage the inherent structure of tensorial data to enhance the modeling capabilities of GPs in the context of complex, spatially varying designs.

Two distinct research directions have recently emerged in the pursuit of integrating GPs with tensorial inputs. The first avenue draws inspiration from convolutional neural networks (CNN). \cite{wilson2016stochastic} proposed a deep kernel learning architecture where the multidimensional inputs were first transformed by a CNN before feeding into the kernel. The large number of parameters introduced by the CNN are treated as kernel hyperparameters. \cite{van2017convolutional} proposed the first convolutional Gaussian process where the process itself was convolved. The same patch-response kernel function was applied to image subpatches and the results were aggregated in the same spirit as additive models. \cite{kumar2018deep} and \cite{blomqvist2020deep} extended the convolutional kernel within a deep GP architecture, further enhancing the model's capacity to capture intricate relationships within image data. The second path has its roots in the tensor regression task. \cite{yu2018tensor} reformulated the low-rank scalar-on-tensor regression problem as a Tensor-GP model with multi-linear kernel. \cite{sun2023tensor} further extended the Tensor-GP model by introducing a dimensionality reduction technique, demonstrating its efficacy on multi-channel imaging data.

While each avenue showcased its capability in capturing local features and multidimensional interactions for tensorial inputs, 
the computational demands associated with these approaches can be prohibitively high, particularly when dealing with tensorial inputs that, while spatially informative, may not possess excessively high dimensions, as is the case with 3D printed antennas. Recognizing the potential overhead incurred by the computationally intensive models, we propose a pragmatic middle-ground strategy. Our approach embeds an Image Distance Metric (IMED) \cite{wang2005euclidean} in replacement of the Euclidean distance in conventional kernels. IMED excels in considering the spatial proximity of pixels/voxels, offering a computationally efficient alternative. The efficacy of our proposed methodology will be illustrated and validated through numerical examples involving 3D printed antennas.

\section{Background}
\label{sec:background}
In this section, we review the Gaussian Process and introduce the Image Euclidean Distance that will lay the groundwork for our proposed method.

\subsection{Notation} 
We represent a voxellized 3D geometric design with associated material properties as a tensor input $\mathcal{X} \in \mathbb{R}^{V\times H\times W\times P}$, where $V$ is the number of voxels along the vertical axis, $H$ is the number of voxels along the horizontal axis, $W$ is the number of voxels along the depth (or width) axis, and $P$ is the number of material properties such as conductivity, permittivity, etc. The material tensor $\mathcal{X}$ can be indexed as $\mathcal{X}^p(i,j,k)$ representing the $p$-th material property at voxel $(i,j,k)$, where 
$1\le i \le V, 1\le j \le H, 1\le W,$ and $1\le p\le P$. The scalar output associated with each input $\mathcal{X}$ is denoted as $y$.

\subsection{Primer on Gaussian processes} 
A Gaussian Process (GP) is a kernel method that characterizes a complete distribution over the function being modeled, \cite{rasmussen2003gaussian} specifies the prior for $y$ as
\begin{equation}
   y = f(\mathcal{X}) + \epsilon, \quad f(\mathcal{X}) \sim \mathcal{GP}(\mu(\mathcal{X}), k(\mathcal{X}, \mathcal{X'})), 
\end{equation}
where $\epsilon$ is the observational noise, $\mu(\mathcal{X})$ is the mean function, and $k$ is the kernel function that measures the similarity between any pair of design inputs $\mathcal{X}$ and $\mathcal{X'}$. Since we are dealing with deterministic computer simulations where \(\epsilon = 0\), to guarantee numerical stability, we adopt a very small value \(\epsilon = 1e-4\) in the model. Additionally, we adopt a constant mean \(\mu(\mathcal{X}) = \mu\) as its sufficiency being proved by extensive studies \cite{ranjan2011computationally,gramacy2015local}.

Consider a 3D geometric dataset with the training data points \(\{\mathcal{X}_i, y_i\}^N_{i=1}\) and the $N_*$ test data points \(\{\mathcal{X}_*, y_*\}\).  Given the inputs $\mathcal{X}$ and 
$\mathcal{X}_*$, the joint distribution of the training output $\mathbf{y}$ and the test output $\mathbf{y}_*$ prior is 
\begin{equation}
\left[ \begin{array}{c}  \mathbf{y}\\ \mathbf{y}_* \end{array} \right] \sim \mathcal{N}(\mu\mathbf{1}, \left[ \begin{array}{cc}  K(\mathbf{\mathcal{X}},\mathbf{\mathcal{X}}) &K(\mathbf{\mathcal{X}},\mathbf{\mathcal{X}_*})\\ K(\mathbf{\mathcal{X}_*},\mathbf{\mathcal{X}}) &
K(\mathbf{\mathcal{X}_*},\mathbf{\mathcal{X}_*})\end{array} \right]),
\end{equation}
where $K(\mathbf{\mathcal{X}},\mathbf{\mathcal{X}_*}) \in \mathbb{R}^{N \times N_*}$ denotes the covariance matrix evaluated at all pairs of training and test points, and similarly for the other entries $K(\mathbf{\mathcal{X}},\mathbf{\mathcal{X}})$, $K(\mathbf{\mathcal{X}_*},\mathbf{\mathcal{X}})$, $K(\mathbf{\mathcal{X}_*},\mathbf{\mathcal{X}_*})$. The posterior distribution is obtained by conditioning the joint Gaussian prior distribution:
\begin{equation}
\begin{split}
\mathbf{y}_*|\mathbf{\mathcal{X}_*},\mathbf{\mathcal{X}},\mathbf{y} \sim &\mathcal{N}(\mu + K(\mathbf{\mathcal{X}_*},\mathbf{\mathcal{X}})K(\mathbf{\mathcal{X}},\mathbf{\mathcal{X}})^{-1}(\mathbf{y}-\mu), \\
&K(\mathbf{\mathcal{X}_*},\mathbf{\mathcal{X}_*}) - K(\mathbf{\mathcal{X}_*},\mathbf{\mathcal{X}})K(\mathbf{\mathcal{X}},\mathbf{\mathcal{X}})^{-1}K(\mathbf{\mathcal{X}},\mathbf{\mathcal{X}}_*)).
\end{split}
\end{equation}

The hyperparameters include $\mu$ and parameters from the kernel function $k$, e.g. lengthscale, signal variance, etc. These parameters are commonly learned by maximizing the log marginal likelihood, given by:
\begin{equation}
    \log p(\mathbf{y}|\mathbf{\mathcal{X}}) = -\frac{1}{2}(\mathbf{y}-\mu)^TK^{-1}(\mathbf{y}-\mu) - \frac{1}{2}\log|K| - \frac{n}{2}\log{2\pi}.
    \label{equ:loglikelihood}
\end{equation}

\subsection{IMED: Image Euclidean Distance} 
The kernel function determines various properties of the GP, such as stationarity, smoothness etc. The most widely used kernel function is the radial basis function (RBF), as it can model any smooth function, given by 
\begin{equation}
    k(\mathcal{X},\mathcal{X}')=\sigma^2 \exp (-\frac{1}{2l}d_E(\mathcal{X},\mathcal{X}')),
\end{equation}
where $\sigma^2$ represents the signal variance and the lengthscale $l$ determines the variations in function values across the inputs. \(d_E(\mathcal{X},\mathcal{X}') = \sum\limits_{p=1}^{P} (vec(\mathcal{X}^p)-vec(\mathcal{X}'^p))^T(vec(\mathcal{X}^p)-vec(\mathcal{X}'^p))\) is the Euclidean distance between two inputs. The assumption underneath the Euclidean distance is that all the base vectors of the input features are orthogonal. One can rewrite the Euclidean distance equation as 
\begin{equation}
    d_E(\mathcal{X},\mathcal{X}') = \sum_{p=1}^{P} (vec(\mathcal{X}^p)-vec(\mathcal{X}'^p))^T G^p (vec(\mathcal{X}^p)-vec(\mathcal{X}'^p)),
\end{equation}
where $G^p \in \mathbb{R}^{VHW \times VHW}$ is the unit diagonal matrix. For a 3D geometric design treated as a data point, the pixel or voxel values serve as features that characterize its configuration. In geometric designs where structural information is inherent; specifically, the pixel or voxel values are not independent but intercorrelated according to their spatial locations, the assumption of an orthogonal feature space is challenged. IMED \cite{wang2005euclidean} was motivated by the fact that images are insensitive to small perturbations which cannot be reflected by traditional Euclidean distance. They propose to accommodate the pixel spatial relationships using non-orthogonal base vectors, i.e., there are off-diagonal elements in matrix $G^p$. To be a valid metric matrix, $G^p$ has to be positive definite which can be fully characterized by a positive definite function, taking Gaussian function as an example:
\begin{equation}
    g^p_{\alpha\beta} = f_p(|J_{\alpha} - J_{\beta}|)=\frac{1}{2\pi(\gamma^p)^2}\text{exp}((-|J_{\alpha} - J_{\beta}|^2/{2(\gamma^p)^2}),
    \label{equ:g_entry}
\end{equation}
where \( g^p_{\alpha\beta} \) represents the element indexed at (\( \alpha, \beta \)) in the matrix \( G^p \) and $\gamma^p$ denotes the lengthscale parameter of material property $p$. $J_{\alpha}$ and $J_{\beta}$ are the $\alpha$th and $\beta$th voxel respectively. Suppose $J_{\alpha}$ is at location $(i,j,k)$, $J_{\beta}$ is at  $(i',j',k')$, then we have:
\[
|J_{\alpha} - J_{\beta}|^2 = (i-i')^2+(j-j')^2+(k-k')^2.
\]

\section{Methodology}
In this section, we introduce the proposed method and discuss its possible extensions, as well as the computational issues.
\subsection{IMED kernel}
\label{sec:IMEDkernel}
For tensor input $\mathcal{X}$ with spatial structures, we propose to embed IMED as the distance measure into existing GP kernels. Taking RBF as an example, incorporating IMED gives
\begin{equation}
    k(\mathcal{X},\mathcal{X}')=\sigma^2 \exp (-\frac{1}{2l}d_{IMED}(\mathcal{X},\mathcal{X}')),
    \label{equ:IMEDkernel}
\end{equation}
\begin{equation}
    d_{IMED}(\mathcal{X},\mathcal{X}') = \sum_{p=1}^{P} (\text{vec}(\mathcal{X}^p)-\text{vec}(\mathcal{X}'^p))^T G^p (\text{vec}(\mathcal{X}^p)-\text{vec}(\mathcal{X}'^p)),
\end{equation}
where $G^p$ is a positive-definite matrix with its entries defined by Equation~\ref{equ:g_entry}. Comparing to the original RBF with $d_E(\mathcal{X},\mathcal{X}')$, only one more hyperparameter $\gamma^p$ is introduced, which governs the locality of pixels/voxels. Direct computation of $d_{IMED}$ using $G^p$ can be expensive. However, leveraging the positive definiteness of $G^p$, the computation can be greatly simplified by introducing a linear transformation. Consider a decomposition of $G^p$, $G^p= {A^p}^TA^p$. If we transform all the tensor input $\mathcal{X}^p$ by $A^p$ and denotes $\mathcal{Z}^p = A^p\text{vec}(\mathcal{X}^p)$, then $d_{IMED}$ between $\mathcal{X}$ and $\mathcal{X}'$ is equal to the traditional Euclidean distance between $\mathcal{Z}$ and $\mathcal{Z}'$, $\mathcal{Z} \in \mathbb{R}^{VHWP}$:
\[
\begin{split}
    d_{IMED}(\mathcal{X},\mathcal{X}') 
    &=\sum_{p=1}^{P} (vec(\mathcal{X}^p)-vec(\mathcal{X}'^p))^T {A^p}^TA^p (vec(\mathcal{X}^p)-vec(\mathcal{X}'^p)) \\
    &=\sum_{p=1}^{P} (\mathcal{Z}^p-\mathcal{Z}'^p)^T  (\mathcal{Z}^p-\mathcal{Z}'^p).
\end{split}
\]
As $G^p$ is solely dependent on the inherent 3D structures, its computation is required only once. Leveraging the transformed tensor input $\mathcal{Z}$ allows us to exploit all the nice properties of conventional Euclidean distance-based kernels. For instance, one can easily introduce the concept of automatic relevance determination (ARD) \cite{bishop2006pattern} into Equation~\ref{equ:IMEDkernel}, which we refer as  ARD-IMED. 

Similarly, the IMED kernel naturally accommodates the directional variations of the 3D structural space by incorporating varying lengthscale $\gamma^p$ associated with different voxel directions in Equation~\ref{equ:g_entry}.

\subsection{Computational issues}
Although the decomposition described in Section~\ref{sec:IMEDkernel} can significantly simplify the computation, the widely used Cholesky decomposition still incurs a computation complexity of $\mathcal{O}((VHW)^3)$, which becomes prohibitive for applications with a large number of voxels. 

An alternative solution was proposed by \cite{gardner2018gpytorch}, wherein the conventional Cholesky decomposition is replaced with the conjugate gradients algorithm. This replacement results in additional reduction of computation time complexity from $\mathcal{O}(N^3)$ to $\mathcal{O}(N^2)$. The utilization of GPU further accelerates the computation. This approach holds promise for addressing the computational challenges associated with large $G^p$.

\section{Simulations and Model Setup}
In this section, we will first introduce the simulation configuration for two numerical examples of 3D-printed antennas. The GP model setup will then be presented. 
\subsection{2D monopole antenna} 
\begin{figure}[h!]
    \centering
    \subfloat[\centering]{{\includegraphics[width=.3\textwidth]{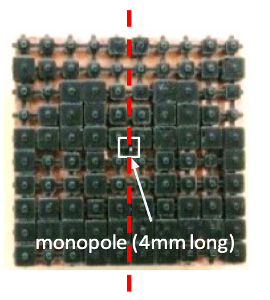} }}%
    % \hspace{1cm}
    \hfill
    \subfloat[\centering]{{\includegraphics[width=.3\textwidth]{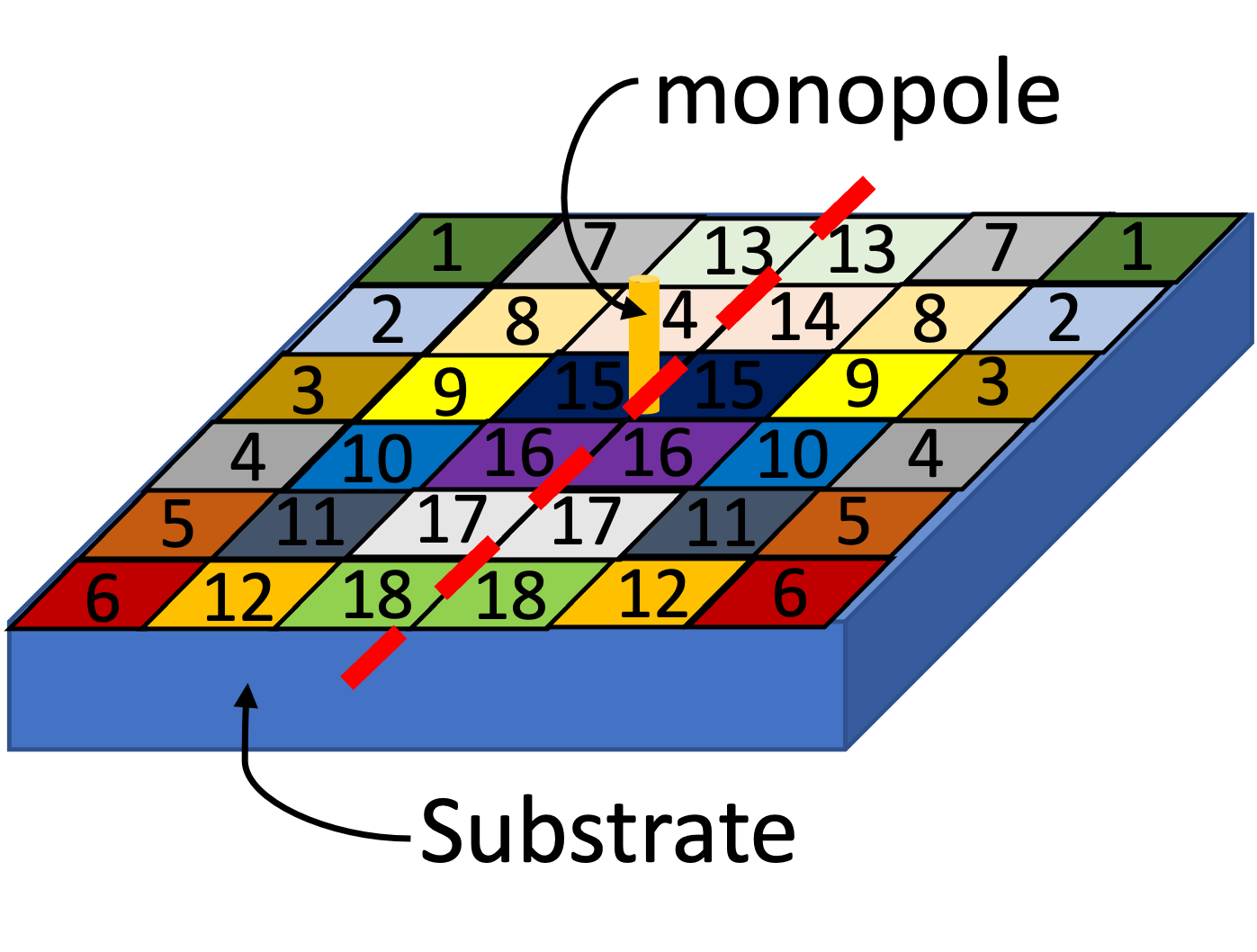} }}%
    % \hspace{1cm}
    \hfill
    \subfloat[\centering]{{\includegraphics[width=.3\textwidth]{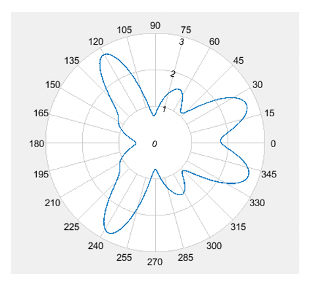} }}%
    \caption[monopole 2D]{2D monopole antenna. (a) a monopole antenna with a 3D-printed dielectric loading, left and right half planes are symmetric. (b) layout of the HFSS model of a monopole antenna surrounded by a grid (6 by 6) of dielectric cells with vertical symmetry. (c) a radiation pattern on a linear scale spanning from 0 to 360 degrees, used as monopole antenna performance measurement.}%
    \label{figchap3:monopole2}%
\end{figure}

A monopole antenna consists of a straight rod-shaped conductor (monopole), which often mounts perpendicularly over a ground plane. Controllable radiation patterns can be achieved by varying the 3D-printed polymer structure with arbitrary dielectric property distribution surrounding the monopole \cite{nayeri20143d}, as shown in Figure~\ref{figchap3:monopole2}(a). In the computer experiment, a quarter-wavelength monopole antenna at the design frequency of 15 GHz was placed on a finite ground plane with size $40\times 40~\mbox{mm}^2$, as shown in Figure~\ref{figchap3:monopole2}(b). The monopole had a 0.5 mm diameter and a 4.8 mm height. It was surrounded by 36 (6 by 6) dielectric unit cells. Each unit cell was designed to have a maximum size of $6.67\times6.67\times6.67~\mbox{mm}$.  Each constant associated with the dielectric property varied continuously from 1.1 to 2.3, which can be physically realized by changing the size of the 3D printed dielectric blocks, similar as in Figure~\ref{figchap3:monopole2}(a). Details of the realization are beyond the scope of this paper; interested readers can refer to \cite{nicolson1970measurement}. The left and right half of the dielectric plane were symmetric (vertically), hence only 18 dielectric constants need to be designed to achieve desired electromagnetic properties. The simulation for the study was configured and executed using Ansys/HFSS. To ensure a representative and space-filling selection of simulation samples, a Latin-hypercube design (LHD) was utilized and 1000 dielectric configurations were sampled.

\subsection{3D monopole antenna} 
\begin{figure}[h!]
    \centering
 {\includegraphics[width=.9\textwidth]{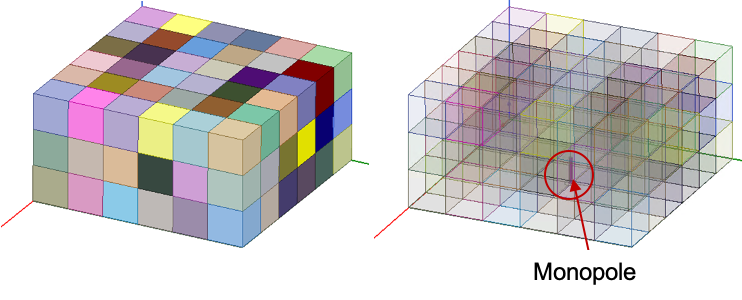} }
    \caption[monopole 3D]{3D monopole antenna. Layout of the HFSS model of a monopole antenna surrounded by $6\times6\times3$ dielectric unit cells }%
    \label{figchap3:monopole3}%
\end{figure}

In the 3D case, the monopole was surrounded by a $6\times6\times3$ dielectric unit cells as shown in Figure~\ref{figchap3:monopole3}. The monopole antenna was located at the center of the ground plane. Each dielectric constant varied continuously from 1.1 to 2.3 as in the 2D examples. The vertical symmetry was removed here, hence all the dielectric constants were considered in the design. We picked 2700 dielectric configurations following LHD.  

\subsection{Functional output} 
The radiation pattern, represented by gain values on a linear scale (refer to Figure~\ref{figchap3:monopole2}(c)), of each dielectric configuration was collected at every angle spanning from 0 to 360 degrees, which yields a functional output $y(\mathcal{X},t), 0\leq t<360$. The B-spline technique was employed for dimension reduction, where a set of B-spline basis vectors were selected for approximation of the original functional output:
\begin{equation}
\left[ \begin{array}{c}  y(\mathcal{X},t_0)\\ \vdots \\ y(\mathcal{X},t_l) \end{array} \right] \approx \sum_{j=1}^{l'}B_{jk}a_j
\end{equation}
where $B_{jk}$ is the $j^{th}$ B-spline of order $k$, and given the order and knot sequence, $B_{jk}$ is completely determined. Therefore, the functional output is uniquely defined by $l'$ number of coefficients $a_j$. In this way, the output dimension is reduced from $l=360$ to $l'$, where $l'$ is often a small number.

For the functional output of the 2D monopole, leveraging its inherent symmetry, only gain values within the range [0, 180] are considered as the model output. After some experimentation, we utilized 21 basis vectors to approximate the radiation curve by splines of order 4, effectively reducing the dimension from 180 to 21. The same order and uniformly placed knot sequences were applied for all the functional outputs. Similarly, the output dimension of the 3D monopole was reduced from 360 to 41. Subsequently, independent GP models were trained on each of the B-spline coefficient $a_j$.
 
After obtaining the predicted mean and variance of all the $a_j$'s, we converted them back to the functional output mean and variance following an independent assumption of the coefficients:
\begin{equation}
\hat{\bm{y}} = \sum_{j=1}^{l'}B_{jk}\hat{a}_j, \quad \hat{\bm{\sigma}}_y^2 = \sum_{j=1}^{l'}B^2_{jk}\hat{\sigma}^2_{a_j}
\label{equ:conversion}
\end{equation}
where $\hat{\bm{y}},\hat{\bm{\sigma}}_y^2 \in \mathbb{R}^l$ are the predicted mean and variance vector of the radiation pattern, respectively. 

\subsection{Model setup and implementations}
For the 2D monopole, the model takes a $6 \times 3$ tensor as input, and the output is a scalar representing the B-spline coefficient. Individual GP models were trained for each coefficient. Once predictions for all coefficients were obtained, we calculated the predicted mean and variance for the radiation pattern at each angle degree following Eq.~\ref{equ:conversion}. For the 3D monopole, the model input becomes a $6 \times 6 \times 3$ tensor. The train/test ratio was 8:2 for both 2D and 3D examples, and both of them satisfy the $10d$ rule of thumb \cite{jones1998efficient} with $d$ being the model input dimension.

We compare the \textbf{IMED} kernel and its variant \textbf{ARD-IMED} to four baseline kernels: \textbf{RBF}, \textbf{ARD-RBF}, the weighted convolutional kernel (details in~\ref{appendix:wconv}), and the multi-linear kernel (\textbf{M-Lin}, obtained by Eq.~\ref{equ:mlinkernel}). For both 2D and 3D monopole designs, we have only one material property (P=1). For the weighted convolutional kernel, two patch sizes were considered for the 2D monopole antenna \{$3 \times 3, 6 \times 3$\} and two cubic sizes for 3D monopole antenna \{$5 \times 5 \times 3, 6 \times 6 \times 3$\}, denoted as \textbf{WConv1} and \textbf{Wconv2}, respectively. ARD-RBF was adopted as the baseline kernel for the patch-response prior $k(z,z')$ in Eq.~\ref{equ:convkernel}. We implemented the IMED, ARD-IMED, WConv1, and WConv2 kernels in the GpyTorch Python package, comparing them to the existing RBF and ARD-RBF functions. The multi-linear (M-Lin) kernel was implemented following the Python script outlined in \cite{sun2023tensor}.

\subsection{Estimation of $G$}

The hyperparameters in the GP with an IMED kernel consist of the kernel lengthscale, signal variance, likelihood noise (constrained to be very small due to the deterministic characteristics of computer simulations) and material property lengthscale $\gamma^p$. We collectively denote this set as $\Theta$, which are commonly learned by minimizing the negative log marginal likelihood 
as computed in Equation~\ref{equ:loglikelihood} and we re-state it here
\begin{equation}
    L(\Theta|\mathbf{\mathcal{X}},\mathbf{y}) \propto -\mathbf{y}^TK^{-1}\mathbf{y} + \log|K|.
    \label{equ:negloglikelihood}
\end{equation}
$\Theta$ is obtained by getting the derivatives
\begin{equation}
    \frac{dL}{d\Theta}=
    \mathbf{y}^TK^{-1}\frac{dK^{-1}}{d\Theta}K^{-1}\mathbf{y} + \text{Tr}(K^{-1}\frac{dK^{-1}}{d\Theta}).
    \label{equ:derivatives}
\end{equation}
An efficient computation scheme for solving Equation~\ref{equ:derivatives} can be found in \cite{gardner2018gpytorch}. After getting the estimate of $\gamma^p$, we can get $\hat G$ as defined in Equation~\ref{equ:g_entry}.

\section{Results Analysis}
In this section, two evaluation metrics will be introduced and then both quantitative and qualitative results will be demonstrated to compare different tensor kernels on the 2D and 3D monopole antenna datasets.
\subsection{Evaluation metrics}
The prediction performance for the radiation pattern was evaluated by using two metrics: the root mean squared error (RMSE) and the mean standard log loss (MSLL)\cite{rasmussen2003gaussian}. The former calculates the rooted average L2 distance of each data point along the radiation pattern curve, and the latter considers the uncertainty quantification. The smaller the MSLL, the more confident and accurate the predictions are. Their formulas are given below:
\begin{equation}
\text{RMSE} = \frac{1}{N_{test}}\sum_{n=1}^{N_{test}}  \sqrt{\frac{\sum_{i=1}^{l}(y_{ni}-\hat{y}_{ni})^2}{l}},
\end{equation}
\begin{equation}
\text{MSLL} = \frac{1}{N_{test} \cdot l}\sum_{n=1}^{N_{test}}  \sum_{i=1}^{l}\left(\frac{1}{2}\text{log}(2\pi\hat{\sigma}_{ni}^2) + \frac{(y_{ni}-\hat{y}_{ni})^2}{2\hat{\sigma}_{ni}^2} \right ),
\end{equation}
where $\hat{y}_{ni}$ represents the mean of the predicted GP at the $i$th angle degree of the $n$th test radiation pattern, and $N_{test}$ is the total number of test samples.

% Coverage probability

\subsection{Quantitative results}
The experiment was iterated 10 times, each with a random train/test set split. Results with a 95\% confidence interval on the test set are summarized in Table~\ref{table:mono2d} for the 2D monopole antenna and Table~\ref{table:mono3d} for the 3D monopole antenna.

For 2D monopole antenna, ARD-IMED achieves statistically significantly smaller RMSE and MSLL, as indicated by a one-sided paired t-test at the significance level $\alpha=0.05$. We can observe a slight improvement after incorporating the spatial correlations introduced by IMED comparing the RBF/IMED and ARD-RBF/ARD-IMED pairs. The weighted convolutional kernel has an error jump when using a smaller patch size $3\times 3$ (WConv1), which can be explained by the fact that we have a rather small input size $6\times 3$. The small receptive field $3\times 3$ struggles to capture complex patterns present in the input. While the patch size $6\times 3$ (WConv2), same as the input size, achieves identical RMSE and slightly smaller MSLL comparing to ARD-RBF. The close results can be explained by the fact that ARD-RBF was used as the base kernel in the patch-response prior. The multi-linear kernel achieves much worse results, which will be further investigated in the qualitative analysis part. Also due to its high computational complexity, only one experiment was conducted.

\begin{table}[h!]
\begin{center}
\caption[2D monopole results comparison]{2D monopole antenna regression performance on the test set under different kernel functions. Results based on 10 random train/test splits and 95\% confidence intervals are reported after $\pm$. WConv1 represents the Weighted Convolutional kernel with the patch size $3\times 3$ and WConv2 uses the patch size $6\times 3$. M-Lin refers the multi-linear kernel. \#params denotes the number of hyperparameters per model.\label{table:mono2d}}
\resizebox{\textwidth}{!}{%
\begin{tabular}{ |c|c|c|c|c|c|c|c| } 
 \hline
 Kernel & RBF & ARD-RBF&IMED&ARD-IMED&WConv1&WConv2&M-Lin \\ 
 \hline
 \hline
 RMSE & 0.099$\pm$0.001 & 0.087$\pm$0.001&0.096$\pm $0.001&\textcolor{red}{0.083$\pm$0.001}&0.373$\pm $0.006&0.087$\pm$0.001 &0.485 \\ 
 \hline
 MSLL & -0.720$\pm$0.022 & -0.876$\pm$0.019&-0.759$\pm$0.025&\textcolor{red}{-0.934$\pm$0.018}&0.546$\pm$0.027&-0.899$\pm$0.018&0.823 \\ 
 \hline
 \#params&3&20&4&21&15&21&45 \\
 \hline
\end{tabular}%
}
\end{center}
\end{table}

For 3D monopole antenna, the results reveals that ARD-IMED has statistically a significantly smaller RMSE while the weighted convolutional kernel with the patch size $6\times 6\times 3$ achieved statistically significantly smaller MSLL, as indicated by a one-sided paired t-test at the significance level $\alpha=0.05$. While all the kernel performance degrades compared to the 2D case as the input dimension increases, the multi-linear kernel achieves better results.

\begin{table}[h!]
\begin{center}
\caption[3D monopole results comparison]{3D monopole antenna regression performance comparison under different kernels. WConv1 represents Weighted Convolutional kernel with cubic size $5\times 5\times 3$ and WConv2 uses cubic size $6\times 6\times 3$.\label{table:mono3d}}
\resizebox{\textwidth}{!}{%
\begin{tabular}{ |c|c|c|c|c|c|c|c| } 
 \hline
 Kernel & RBF & ARD-RBF&IMED&ARD-IMED&WConv1&WConv2&M-Lin \\ 
 \hline
 \hline
 RMSE & 0.306$\pm$0.002 & 0.275$\pm$0.002&0.299$\pm$0.001&\textcolor{red}{0.267$\pm$0.001}&0.645$\pm$0.007&0.270$\pm$0.002&0.455 \\ 
 \hline
 MSLL & 0.408$\pm$0.004 & 0.318$\pm$0.004&0.380$\pm$0.004&0.284$\pm$0.004&0.927$\pm$0.011&\textcolor{red}{0.237$\pm$0.005}&0.736 \\ 
 \hline
 \#params&3&110&4&111&81&111&81 \\
 \hline
\end{tabular}%
}
\end{center}
\end{table}

\subsection{Qualitative analysis}
\begin{figure}[h!]
    \centering
    \subfloat[\centering]{{\includegraphics[width=.45\textwidth]{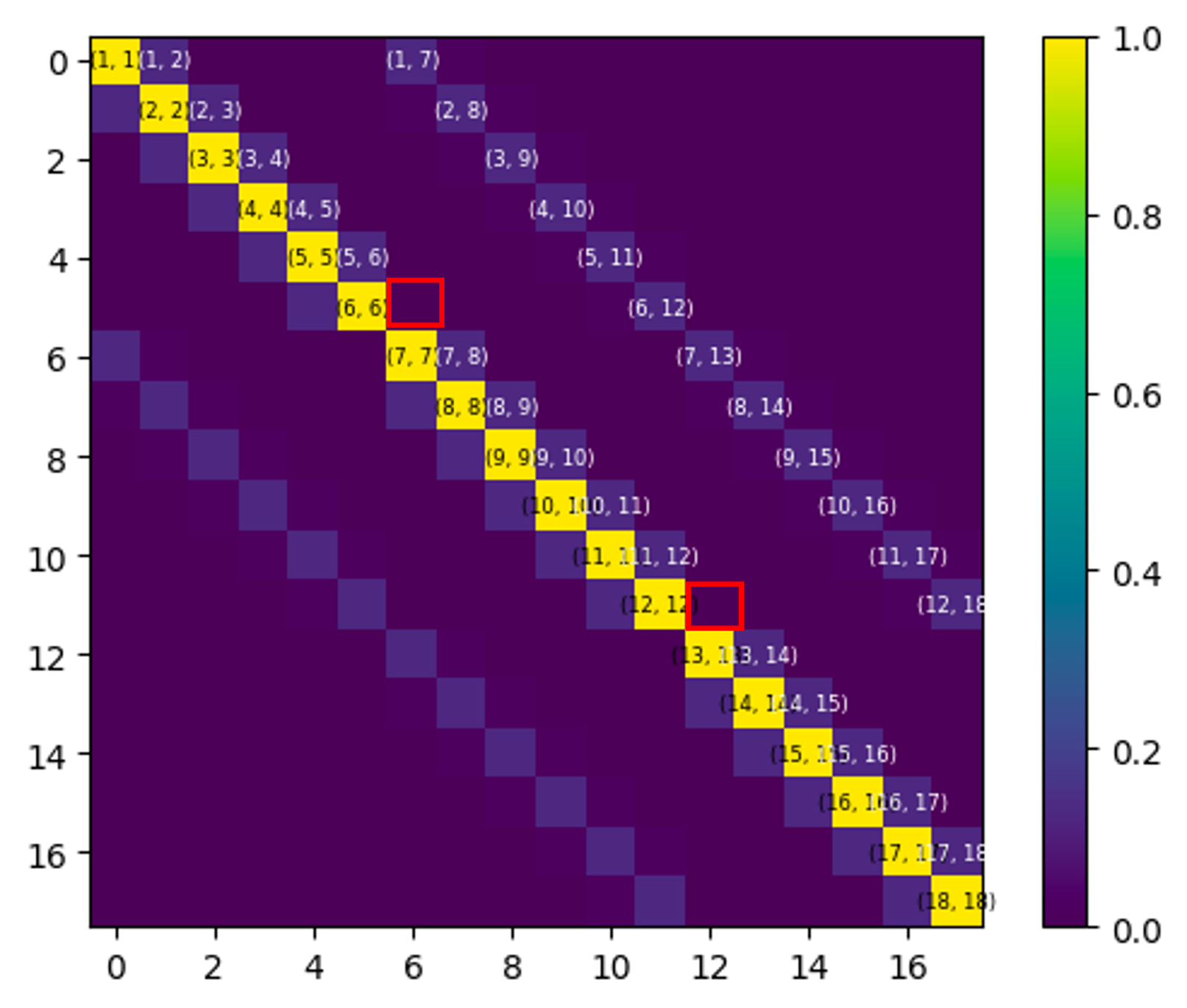} }}%
    % \hspace{1cm}
    \hfill
    \subfloat[\centering]{{\includegraphics[width=.45\textwidth]{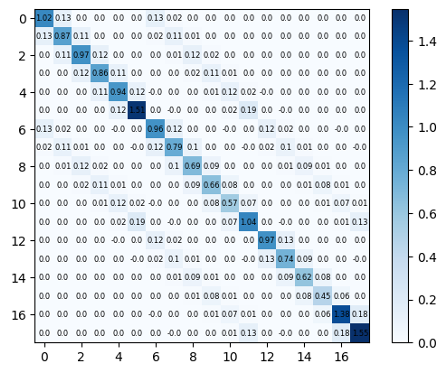} }}%
    \hfill
    \subfloat[\centering]{{\includegraphics[width=.45\textwidth]{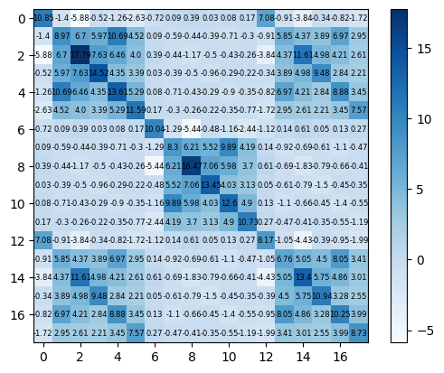} }}%
    \hfill
    \subfloat[\centering]{{\includegraphics[width=.45\textwidth]{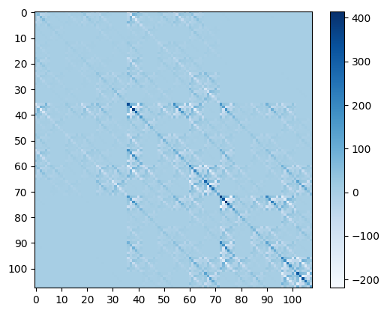} }}%
    \caption[Kernel check]{(a) one randomly selected estimated matrix $\hat{G}$ for 2D monopole, off-diagonal elements account for correlations among pixels whose indices are annotated. (b) an estimated $\hat{G}$ in ARD-IMED for 2D monopole with the pixel value annotated. (c) an estimated $\hat{K}=\hat{K}_2\otimes\hat{K}_1$ in M-Lin for 2D monopole with the pixel value annotated. (d) an estimated $\hat{K}=\hat{K}_3\otimes\hat{K}_2\otimes\hat{K}_1$ in M-Lin for 3D monopole.}%
    \label{figchap3:Gmatrix2D}%
\end{figure}

We investigate the working mechanism of IMED and M-Lin. For simplicity, we choose the 2D monopole for illustration. The magic of IMED lies in the distance metric matrix $G$ which gauges the proximity of features by considering pixel closeness. A randomly chosen estimated matrix $\hat{G}$ is displayed in Figure~\ref{figchap3:Gmatrix2D}(a), showcasing correlations among neighboring pixels through off-diagonal elements. The two red squares on the matrix, highlighting missing (extremely small) correlations, correspond to indices (6,7) and (12,13). These indices align with the dielectric cells in Figure~\ref{figchap3:monopole2}(b), emphasizing that the 6th and 7th dielectric cells, as well as the 12th and 13th dielectric cells, are actually spatially distant from each other. 

Figure~\ref{figchap3:Gmatrix2D}(b) shows the estimated $\hat{G}$ in ARD-IMED where different lengthscales were considered for each dielectric cell. The visualization of the estimated $\hat{K}$ matrix of the M-LIN kernel shown in Figure~\ref{figchap3:Gmatrix2D}(c) unveils its essence, where pixel correlations are defined as the product of column and row similarities. Although the performance was suboptimal in the 2D monopole case, a substantial improvement was observed in the 3D monopole scenario. Examination of the estimated $\hat{K}$ in Figure~\ref{figchap3:Gmatrix2D}(d) reveals a structure much closer to the ARD-IMED kernel, providing evidence that IMED effectively captured the underlying structural information.

\begin{figure}[h!]
    \centering
    {\includegraphics[width=\textwidth]{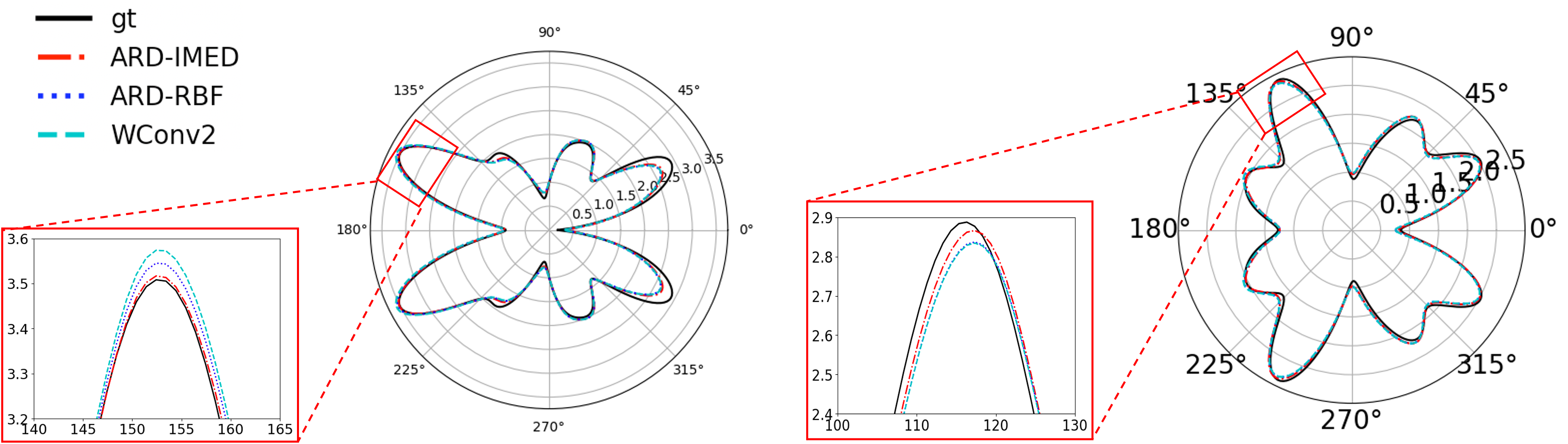} }%
    % \hspace{1cm}
    \caption[main lobe]{Predicted mean for radiation pattern at the main lobe of randomly selected 2D monopole designs, ARD-IMED outperforms ARD-RBF and WConv2.}%
    \label{figchap3:mainlobe}%
\end{figure}

In the context of radiation patterns, the main lobe represents the primary direction where the antenna emits the majority of its energy. Predicting this region poses a considerable challenge due to its steep nature. We define the main lobe as the $\pm 7$ degrees around the peaks of each radiation pattern, constituting approximately 11\% of the total squared error in the entire pattern. When recalculating the RMSE and MSLL exclusively for the main lobe areas of the 2D monopole, as detailed in Table~\ref{table:lobe2d}, notable increases in MSLL values were observed across all compared models, indicating a rise in prediction uncertainty. Notably, ARD-IMED consistently outperforms ARD-RBF and WConv2 under both evaluation metrics. The nuances of these kernels are further illustrated by two randomly selected sample predictions in Figure~\ref{figchap3:mainlobe}.
\begin{table}[h!]
\begin{center}
\caption[2D monopole main lobe results comparison]{2D monopole antenna regression performance comparison at the main lobe. \label{table:lobe2d}}
\resizebox{0.6\textwidth}{!}{%
\begin{tabular}{|c|c|c|c|} 
 \hline
 Kernel & ARD-RBF & ARD-IMED & WConv2 \\ 
 \hline
 \hline
 RMSE & 0.095$\pm$0.002 & \textcolor{red}{0.090$\pm$0.002} & 0.093$\pm$0.002 \\ 
 \hline
 MSLL & -0.362$\pm$0.073 & \textcolor{red}{-0.435$\pm$0.07} & -0.423$\pm$0.066 \\ 
 \hline
\end{tabular}%
}
\end{center}
\end{table}

\section{Conclusion}
In this paper, we propose to integrate IMED into the Gaussian process, allowing for learning structural information for 3D printed antennas in a supervised learning context. We see improvements on the regression performance over the previous convolutional kernel and multi-linear kernel in both 2D and 3D monopole tasks. The capability of the model in generating an interpretable structural representation makes it ideal for spatially structured data with moderately sized dimensions. The current model has several limitations. First, the computational complexity introduced by the metric matrix $G$ poses challenges when dealing with higher-dimensional data. Exploring techniques such as the conjugate gradient algorithm \cite{gardner2018gpytorch} could offer promising solutions. Second, while IMED essentially serves as a transformation function for inputs, excelling in its simplicity and interpretability with the introduction of only one additional hyperparameter, it may lack the flexibility to capture more intricate structural information. On the other hand, neural network-based transformation functions provide greater flexibility but come at the cost of increased computational complexity. A middle ground should be found.

\appendix
% \section{Appendix}

\section{Weighted Convolutional kernel}
\label{appendix:wconv}
The convolutional kernel was inspired directly from the convolutional neural network (CNN) where the function evaluation on an image is considered as the sum of functions over the patches of the input image \cite{van2017convolutional}. Given the 3D geometric input $\mathcal{X} \in \mathbb{R}^{V\times H\times W\times P}$, we introduce the same idea of the multi-channel convolutional kernel, where we define patch-response functions $g_p: \mathbb{R}^{v \times h \times w} \rightarrow \mathbb{R}$ for each material property $p$ to take a $v \times h \times w$ patch input. We obtain a total of $M=(V-v+1)\times(H-h+1)\times(W-w+1)$ patches. The overall function is obtained by summing over all the patch responses. If giving $g_p(\cdot)$ a GP prior, we can still get a GP prior on $f(\cdot)$:
\begin{equation}
   g_p \sim \mathcal{GP}(0, k_p(z,z')),
   \quad
   f(\mathcal{X}) = \sum_{m=1}^{M}\sum_{p=1}^{P}w_{mp}g_p(\mathcal{X}^{[mp]})
\end{equation}
\begin{equation}
   \Rightarrow f(\mathcal{X}) \sim \mathcal{GP}(0, \sum_{m=1}^{M}\sum_{m'=1}^{M}(\sum_{p=1}^{P}\sum_{p'=1}^{P}w_{mp}w_{m'p'}k_p(\mathcal{X}^{[mp]},\mathcal{X'}^{[mp]'})))
   \label{equ:convkernel}
\end{equation}
where $\mathcal{X}^{[mp]}$ indicates the $m$th patch for the $p$th material of $\mathcal{X}$, the weights $\{w_{mp}\}$ adjust the relative importance of the response for each material property at each location of the 3D geometry.

Implementation of the convolutional kernel requires $(PM)^2$ evaluations for each entry of $K_{\mathcal{X},\mathcal{X'}}$.

\section{Multi-linear kernel}
\label{appendix:mlin}
The multi-linear tensor kernel is given by:
\begin{equation}
k(\mathcal{X},\mathcal{X'}) = \text{vec}(\mathcal{X})^T(K_4 \otimes K_3 \otimes K_2 \otimes K_1)\text{vec}(\mathcal{X'})
\label{equ:mlinkernel}
\end{equation}
where $\text{vec}(\cdot)$ is the vectorization operator and $\otimes$ denotes the matrix Kronecker product. $K_1 \in \mathbb{R}^{V\times V}, K_2 \in \mathbb{R}^{H\times H}, K_3 \in \mathbb{R}^{W\times W}, K_4 \in \mathbb{R}^{P\times P}$ captures the mode-specific covariance structure. 

To speed up the computation, each multi-linear kernel factor can be approximated with a factorized form:
\begin{equation}
   K_1=U_1^TU_1,
   \quad
   K_2=U_2^TU_2
   \quad
   K_3=U_3^TU_3
   \quad
   K_4=U_4^TU_4
\end{equation}
where $U_1 \in \mathbb{R}^{v\times V}, U_2 \in \mathbb{R}^{h\times H}, U_3 \in \mathbb{R}^{w\times W}, U_4 \in \mathbb{R}^{p\times P}$. $U_1, U_2, U_3, U_4$ are orthogonal matrices with $v \leq V, h \leq H, w \leq W, p \leq P$. The tuning parameter is set as such that $v = V, h = H, w = W, p = P$ throughout the paper but can be set to smaller values to enforce a low-rank constraint. With the factorization assumption, one can decompose the gram matrix $K$ as $\Tilde{U}\Tilde{U}^T$, where:
\begin{equation}
\Tilde{U} = \Tilde{\mathcal{X}}^T(U_4 \otimes U_3 \otimes U_2 \otimes U_1)^T
\end{equation}
where $\mathcal{X}=[\text{vec}(\mathcal{X}_1);\text{vec}(\mathcal{X}_2);\dots;\text{vec}(\mathcal{X}_N)]$. The factorized form of $K$ can reduce the computational complexity from $\mathcal{O}(N^3)$ to $\mathcal{O}(N^2D)$, where $D=VHWP$ is the dimension of the data tensor.

% \section*{Acknowledgments}
% This was was supported in part by......

%Bibliography
\bibliographystyle{unsrt}  
\bibliography{references}

\begin{thebibliography}{10}

\bibitem{conti2010bayesian}
Stefano Conti and Anthony O’Hagan.
\newblock Bayesian emulation of complex multi-output and dynamic computer models.
\newblock {\em Journal of statistical planning and inference}, 140(3):640--651, 2010.

\bibitem{fricker2013multivariate}
Thomas~E Fricker, Jeremy~E Oakley, and Nathan~M Urban.
\newblock Multivariate gaussian process emulators with nonseparable covariance structures.
\newblock {\em Technometrics}, 55(1):47--56, 2013.

\bibitem{koziel2013surrogate}
Slawomir Koziel and Leifur Leifsson.
\newblock {\em Surrogate-based modeling and optimization}.
\newblock Springer, 2013.

\bibitem{rasmussen2003gaussian}
Carl~Edward Rasmussen.
\newblock Gaussian processes in machine learning.
\newblock In {\em Summer school on machine learning}, pages 63--71. Springer, 2003.

\bibitem{santner2003design}
Thomas~J Santner, Brian~J Williams, William~I Notz, and Brain~J Williams.
\newblock {\em The design and analysis of computer experiments}, volume~1.
\newblock Springer, 2003.

\bibitem{wang2022bayesian}
Kunbo Wang and Yanxun Xu.
\newblock Bayesian tensor-on-tensor regression with efficient computation.
\newblock {\em arXiv preprint arXiv:2210.11363}, 2022.

\bibitem{wilson2016stochastic}
Andrew~G Wilson, Zhiting Hu, Russ~R Salakhutdinov, and Eric~P Xing.
\newblock Stochastic variational deep kernel learning.
\newblock {\em Advances in neural information processing systems}, 29, 2016.

\bibitem{van2017convolutional}
Mark Van~der Wilk, Carl~Edward Rasmussen, and James Hensman.
\newblock Convolutional gaussian processes.
\newblock {\em Advances in Neural Information Processing Systems}, 30, 2017.

\bibitem{kumar2018deep}
Vinayak Kumar, Vaibhav Singh, PK~Srijith, and Andreas Damianou.
\newblock Deep gaussian processes with convolutional kernels.
\newblock {\em arXiv preprint arXiv:1806.01655}, 2018.

\bibitem{blomqvist2020deep}
Kenneth Blomqvist, Samuel Kaski, and Markus Heinonen.
\newblock Deep convolutional gaussian processes.
\newblock In {\em Machine Learning and Knowledge Discovery in Databases: European Conference, ECML PKDD 2019, W{\"u}rzburg, Germany, September 16--20, 2019, Proceedings, Part II}, pages 582--597. Springer, 2020.

\bibitem{yu2018tensor}
Rose Yu, Guangyu Li, and Yan Liu.
\newblock Tensor regression meets gaussian processes.
\newblock In {\em International Conference on Artificial Intelligence and Statistics}, pages 482--490. PMLR, 2018.

\bibitem{sun2023tensor}
Hu~Sun, Ward Manchester, Meng Jin, Yang Liu, and Yang Chen.
\newblock Tensor gaussian process with contraction for multi-channel imaging analysis.
\newblock {\em arXiv preprint arXiv:2301.11203}, 2023.

\bibitem{wang2005euclidean}
Liwei Wang, Yan Zhang, and Jufu Feng.
\newblock On the euclidean distance of images.
\newblock {\em IEEE transactions on pattern analysis and machine intelligence}, 27(8):1334--1339, 2005.

\bibitem{ranjan2011computationally}
Pritam Ranjan, Ronald Haynes, and Richard Karsten.
\newblock A computationally stable approach to gaussian process interpolation of deterministic computer simulation data.
\newblock {\em Technometrics}, 53(4):366--378, 2011.

\bibitem{gramacy2015local}
Robert~B Gramacy and Daniel~W Apley.
\newblock Local gaussian process approximation for large computer experiments.
\newblock {\em Journal of Computational and Graphical Statistics}, 24(2):561--578, 2015.

\bibitem{bishop2006pattern}
Christopher~M Bishop and Nasser~M Nasrabadi.
\newblock {\em Pattern recognition and machine learning}, volume~4.
\newblock Springer, 2006.

\bibitem{gardner2018gpytorch}
Jacob Gardner, Geoff Pleiss, Kilian~Q Weinberger, David Bindel, and Andrew~G Wilson.
\newblock Gpytorch: Blackbox matrix-matrix gaussian process inference with gpu acceleration.
\newblock {\em Advances in neural information processing systems}, 31, 2018.

\bibitem{nayeri20143d}
Payam Nayeri, Min Liang, Rafael~Austreberto Sabory-Garc{\i}, Mingguang Tuo, Fan Yang, Michael Gehm, Hao Xin, Atef~Z Elsherbeni, et~al.
\newblock 3d printed dielectric reflectarrays: Low-cost high-gain antennas at sub-millimeter waves.
\newblock {\em IEEE Transactions on Antennas and Propagation}, 62(4):2000--2008, 2014.

\bibitem{nicolson1970measurement}
AM~Nicolson and GF~Ross.
\newblock Measurement of the intrinsic properties of materials by time-domain techniques.
\newblock {\em IEEE Transactions on instrumentation and measurement}, 19(4):377--382, 1970.

\bibitem{jones1998efficient}
Donald~R Jones, Matthias Schonlau, and William~J Welch.
\newblock Efficient global optimization of expensive black-box functions.
\newblock {\em Journal of Global optimization}, 13:455--492, 1998.

\end{thebibliography}

\end{document}